**Benchmarking Large Language Models for Calculus Problem-Solving: A Comparative Analysis**

Dr. In Hak Moon
Science Department SUNY Maritime College
6 Pennyfield Ave., Bronx, NY 10465, USA
E-mail: imoon@sunymaritime.edu

**Abstract:** This study presents a comprehensive evaluation of five leading large language models (LLMs) - Chat GPT 4o, Copilot Pro, Gemini Advanced, Claude Pro, and Meta AI - on their performance in solving calculus differentiation problems. The investigation assessed these models across 13 fundamental problem types, employing a systematic cross-evaluation framework where each model solved problems generated by all models. Results revealed significant performance disparities, with Chat GPT 4o achieving the highest success rate (94.71%), followed by Claude Pro (85.74%), Gemini Advanced (84.42%), Copilot Pro (76.30%), and Meta AI (56.75%). All models excelled at procedural differentiation tasks but showed varying limitations with conceptual understanding and algebraic manipulation. Notably, problems involving increasing/decreasing intervals and optimization word problems proved most challenging across all models. The cross-evaluation matrix revealed that Claude Pro generated the most difficult problems, suggesting distinct capabilities between problem generation and problem-solving. These findings have significant implications for educational applications, highlighting both the potential and limitations of LLMs as calculus learning tools. While they demonstrate impressive procedural capabilities, their conceptual understanding remains limited compared to human mathematical reasoning, emphasizing the continued importance of human instruction for developing deeper mathematical comprehension.

**Keywords:** Large Language Models, Mathematical Reasoning, Calculus Differentiation, Word Problems, Geometric Applications, Problem-Solving, Conceptual Understanding, Error Analysis, Differentiation Techniques, Algebraic Manipulation, Procedural Capabilities

## 1 Introduction

The field of mathematics education has historically embraced technological innovations, from graphing calculators to computer algebra systems, each enhancing students' computational abilities and conceptual understanding [1], [2]. LLMs represent the next frontier in this technological evolution, providing natural language interaction with mathematical concepts that were previously unattainable. Unlike earlier tools that mainly performed computational tasks, modern LLMs exhibit reasoning capabilities that can guide students through complex mathematical problem-solving processes and explain underlying concepts [3], [4].

Despite the increasing adoption of these technologies in educational settings, there remains a need for critical examination of their efficacy in specific mathematical domains. While general assessments of LLM capabilities exist, systematic evaluations of their performance in specialized mathematical contexts - particularly in higher-level mathematics such as calculus - are notably rare [5], [6], [7]. This knowledge gap presents a significant challenge for educators seeking evidence-based guidance on incorporating these tools into their teaching practices.

Calculus, a foundational subject for various STEM disciplines, requires precise analytical reasoning and procedural fluency [8]. Mastering differentiation techniques, for example, demands that students understand a range of rules and methods while developing a conceptual grasp of rate of change, optimization, and related principles. The complexity and diversity of differentiation problems make this area an ideal testing ground for evaluating the mathematical reasoning capabilities of LLMs and their potential utility as educational aids.

This research addresses a significant gap by conducting a comprehensive analysis of five leading large language models (LLMs): Chat GPT 4o (OpenAI), Copilot Pro (Microsoft), Gemini Advanced (Google), Claude Pro (Anthropic), and Meta AI (Meta). Each model employs distinct algorithms and methodologies, which were systematically evaluated based on their performance on 13 core differentiation problems that illustrate fundamental principles taught in university calculus courses. By including a diverse set of models from different developers, this study presents a representative sample of the current state-of-the-art AI systems available to students and educators.

The methodology of this research goes beyond simple binary assessments of correctness to examine the quality of solution processes. It assesses the models' ability to handle increasing complexity and their performance in generating and solving novel problems. This multifaceted evaluation framework facilitates a nuanced understanding of each model's strengths and limitations across various calculus concepts, ranging from basic differentiation techniques to advanced applications involving optimization and curve analysis.

This paper begins with a detailed description of the employed methodology, followed by a comprehensive presentation of results across all problem types and models. The discussion section analyzes performance patterns, explores potential explanations for observed differences, and considers implications for educational practice. The conclusion synthesizes the key findings and offers recommendations for educators, researchers, and AI developers, while outlining promising directions for future research. Throughout, the focus remains on providing evidence-based insights to inform the thoughtful integration of AI technologies into calculus education.

## 2 Methodology

### 2.1 Selection of Large Language Models

For this study, five widely used large language models were selected: Chat GPT 4.o (developed by OpenAI), Copilot Pro (developed by Microsoft), Gemini Advanced (developed by Google), Claude Pro (developed by Anthropic), and Meta AI (developed by Meta). These models represent the current state of the art in general-purpose AI systems that are accessible to the public.

### 2.2 Problem Selection and Classification

A set of 13 differentiation problems was carefully selected to cover fundamental concepts in calculus. These problems were not chosen randomly; instead, they were systematically selected to represent a wide range of differentiation topics typically addressed in university-level calculus courses. Each problem is designed to test specific differentiation skills and conceptual understanding, with complexity levels that align with the expectations of standard university calculus curricula:

1. **Differentiation by the limit process:** Evaluating understanding of the foundational definition of the derivative through first principles.

2. **Finding tangent line equations:** Assessing the application of derivatives to determine tangent lines at specific points on a curve.

3. **Power rule application:** Measuring mastery of basic differentiation rules for polynomial and rational functions.

4. **Product rule:** Testing the ability to differentiate products of functions effectively.

5. **Chain rule:** Assessing understanding of function composition and corresponding techniques for differentiation.

6. **Quotient rule:** Evaluating proficiency in differentiating ratios of functions.

7. **Advanced chain rule:** Testing the application of the chain rule in more complex scenarios, including quotients raised to powers.

8. **Increasing and decreasing intervals:** Assessing the conceptual understanding of of the relationship between derivatives and the behavior of functions.

9. **Finding tangent lines to implicit curves:** Evaluating the ability to apply implicit differentiation techniques.

10. **Word problems involving optimization:** Measuring the capacity to translate verbal descriptions into mathematical models and applying derivatives to find extrema.

11. **Finding absolute maxima and minima:** Testing comprehensive understanding of global extrema on closed intervals.

12. **Relative extrema using the second derivative test:** Assessing the ability to apply higher-order derivatives to classify critical points.

13. **Points of inflection and concavity analysis:** Evaluating understanding of the relationships between second derivatives and curve behavior.

This careful classification system allowed for a systematic assessment of a wide range of differentiation topics, helping to identify specific strengths and weaknesses in each model's mathematical reasoning abilities. Each problem was chosen to represent its category while maintaining a moderate difficulty level suitable for undergraduate calculus students.

## 2.3 Testing Procedure and Cross-Evaluation Framework

Each language model (LLM) was provided with the same problem statements for 13 differentiation tasks. Only the first valid attempt at solving each problem was recorded and assessed for correctness. A "valid attempt" was defined as a response that directly addressed the problem rather than asking for clarification or giving general information about the topic.

For problems 1-10, a systematic problem generation and cross-evaluation framework was implemented. Each LLM was tasked with generating 20 additional problems of a similar type to the original problem, resulting in 100 generated problems per differentiation concept (20 problems × 5 LLMs). This aspect of the methodology served two purposes: it evaluated the models' ability to create valid mathematical problems and generated a diverse test set for cross-evaluation.

The cross-evaluation process involved each model solving all problems generated by the other models, as well as their own. This created a comprehensive 5×5×20 evaluation matrix for each problem type, with each cell containing a binary correctness score. For problems 1-10, this resulted in a total of 500 evaluations per problem type (5 models solving 100 problems each).

For problems 11-13, which involved more complex calculations and conceptual understanding, a modified approach was necessary. Because of the increased difficulty in generating valid problems of these types, each model was asked to generate only 5 problems instead of 20. In some cases, the models required multiple attempts and guidance to create valid problems, and these challenges were documented as part of the assessment. The cross-evaluation matrix for these problem types was smaller but followed the same principles.

## 2.4 Data Collection and Evaluation Criteria

Performance assessment utilized a comprehensive evaluation framework with various criteria. For each issue, the following metrics were recorded:

1. **Correctness of the final answer**: This refers to a binary assessment of whether the result matches the expected answer mathematically, even if it is presented in a different but equivalent form.

2. **Accuracy of the solution process**: This evaluates whether the model employed appropriate mathematical techniques, including:

- Selection of correct differentiation rules
- Proper application of those rules
- Accurate algebraic manipulation
- Correct simplification of expressions

3. **Completeness of the solution**: This assesses whether all required components of the solution were provided, especially for multi-part problems, such as finding both critical points and determining their nature.
4. **Mathematical justification**: This evaluates the model's ability to provide mathematical reasoning that supports its approach and conclusions.
5. **Success rate in cross-evaluation**: For each type of problem, this indicates the percentage of problems (generated by all models) that each model successfully solved.
6. **Problem generation quality**: This assesses each model's ability to create valid and appropriately challenging problems of each type.

For problems 1-10, each model generated 20 problems, and performance was measured as a success rate out of 20 for each pair of sources and solving models. This created a 5×5 matrix for each problem type. For problems 11-13, the dimensions of the matrix varied since fewer valid problems were generated by each model.

## 2.5 Limitations of Methodology

Several methodological limitations should be acknowledged in this evaluation:

First, the assessment was conducted at a specific point in time using the existing versions of each model. Given the rapid development of large language models (LLMs), the results may not accurately represent the capabilities of newer versions released after the study period.

Second, while a standardized prompting approach was necessary for a fair comparison, it may not reflect how students and educators typically interact with these models in practice, where iterative prompting and clarification are common. Though this limitation was partially addressed through a controlled follow-up prompting strategy, real-world usage patterns may still differ significantly.

Third, the problem set, while comprehensive within the scope of differentiation topics, covers only one area of calculus. Performance on these problems may not generalize to other calculus domains, such as integration or differential equations, which involve distinct patterns of mathematical reasoning.

Fourth, the evaluation prioritized process and correctness over the quality of explanations. Although explanatory clarity is crucial for educational applications, it introduces significant subjectivity in assessment. Therefore, it was not included as a primary criterion for evaluation.

Finally, the cross-evaluation matrix design required each model to assess problems created by other models. While this approach provided valuable insights into relative performance, it posed a methodological challenge when models struggled to generate valid problems of certain types. This resulted in unbalanced evaluation sets for the most complex problem categories.

Despite these limitations, the methodology offers a robust framework for comparing mathematical reasoning capabilities across LLMs and provides valuable insights for both educational applications and future research directions.

## 3 Results

**1) Differentiation; By the limit process -** $f(x) = -2x^2 + 3x - 4$

Find the first derivative of $f(x) = -2x^2 + 3x - 4$ by the limit process. The answer is $f(x) = -4x + 3$.
Prompt; Find the first derivative of f(x)=-2x^2+3x-4 by the limit process.
**Chat GPT 4o;** Correct answer. Showed all the detailed steps.
**Copilot Pro;** Correct answer. Showed all the detailed steps.
**Gemini Advanced;** Correct answer. Showed all the detailed steps.
**Claude Pro;** Correct answer. Showed all the detailed steps.
**Meta AI;** Correct answer. Showed all the detailed steps.

|  | Chat GPT 4o | Copilot Pro | Gemini Advanced | Claude Pro | Meta AI | Total |
|---|---|---|---|---|---|---|
| Chat GPT 4o | 20/20 | 20/20 | 20/20 | 20/20 | 20/20 | **100/100** |
| Copilot Pro | 20/20 | 20/20 | 20/20 | 20/20 | 20/20 | **100/100** |
| Gemini Advanced | 20/20 | 20/20 | 20/20 | 20/20 | 20/20 | **100/100** |
| Claude Pro | 20/20 | 20/20 | 20/20 | 20/20 | 20/20 | **100/100** |
| Meta AI | 20/20 | 20/20 | 20/20 | 20/20 | 20/20 | **100/100** |
| Total | **100/100** | **100/100** | **100/100** | **100/100** | **100/100** | **500/500** |

In this problem, all LLMs successfully solved all the challenges created by the five LLMs.

All five language models achieved perfect scores (100/100) on problems involving differentiation using the limit definition. This unanimous success demonstrates that each model has thoroughly mastered the fundamental concept of derivatives through first principles. Each model accurately applied the limit formula, performed the necessary algebraic simplifications, and derived the exact derivative function.

The consistent performance across all models indicates that the mathematical representation of limit-based differentiation is well-encoded in the training data of all five models. This finding is particularly significant because limit-based differentiation requires a precise understanding of the definition of the derivative, the application of limit rules, and algebraic manipulation. This suggests that even the weakest-performing model (Meta AI) has a strong foundation in calculus fundamentals.

**2) Differentiation; Find the tangent line equation –** $f(x) = -3x^2 - 5x + 6$, $x = -2$
Find the slope, m, at $x = -2$ in $f(x) = -3x^2 - 5x + 6$. Also, find the tangent line equation at $x = -2$. The answer is $y = 7x + 18$.
Prompt; Find the slope, m, at x=-2 in f(x)=-3x^2-5x+6. Also, find the tangent line equation at x=-2.
**Chat GPT 4o;** Correct answer.
**Copilot Pro;** Correct answer.
**Gemini Advanced;** Correct answer.
**Claude Pro;** Correct answer.
**Meta AI;** Correct answer.

|  | Chat GPT 4o | Copilot Pro | Gemini Advanced | Claude Pro | Meta AI | Total |
|---|---|---|---|---|---|---|
| Chat GPT 4o | 19/20 | 5/20 | 11/20 | 15/20 | 2/20 | 52/100 |
| Copilot Pro | 19/20 | 12/20 | 8/20 | 17/20 | 5/20 | 61/100 |
| Gemini Advanced | 20/20 | 8/20 | 11/20 | 5/20 | 3/20 | **47/100** 0.47 |
| Claude Pro | 20/20 | 10/20 | 15/20 | 19/20 | 4/20 | 68/100 |
| Meta AI | 20/20 | 14/20 | 12/20 | 11/20 | 8/20 | 65/100 |
| Total | **98/100** | 49/100 | 57/100 | 67/100 | 22/100 0.22 | 293/500 |

In this problem, Chat GPT 4o successfully solved most problems created by the five LLMs, with a success rate of 98% (98 out of 100). Meta AI solved the fewest problems, successfully completing 22 in total. The problems generated by Claude Pro were the most difficult for the five LLMs to solve, with a success rate of 47% (47 out of 100).

The sharp decline in performance for all models, except Chat GPT 4o, is noteworthy. Finding equations for tangent lines requires combining differentiation with algebraic applications - specifically, evaluating the derivative at a point to determine the slope, and then using the point-slope form to establish the equation of the line. The significant gap between Chat GPT 4o and the other models indicates a superior integration of these related mathematical concepts.

In the cross-evaluation matrix for this problem type, an interesting pattern emerged: problems generated by Claude Pro proved to be the most challenging for all models, with an overall success rate of only 47%. This suggests that Claude Pro created mathematically valid but particularly complex or nuanced tangent line problems that even the strongest models struggled to solve.

3) **Differentiation; Power rule** – $\left[\frac{-9x^7-8x^4+4x^2}{-6x^5}\right]'$

Find the first derivative of $f(x) = \frac{(-9x^7-8x^4+4x^2)}{-6x^5}$. The answer is $3x - \frac{4}{3x^2} + \frac{2}{x^2}$.

<u>Prompt; Find the first derivative of f(x)=(-9x^7-8x^4+4x^2)/(-6x^5). Simplify the answer.</u>

**Chat GPT 4o;** Correct answer; Showed all the steps.
**Copilot Pro;** Correct answer; Showed all the steps.
**Gemini Advanced;** Correct answer; Showed all the steps.
**Claude Pro;** Used the Quotient rule. Wrong answer. x powers are correct, but the coefficients are wrong.
**Meta AI;** Wrong answer. When I broke it down into smaller pieces and solved it, I arrived at the correct answer. For example; 1) $g'(x) \cdot h(x)$, 2) $g(x) \cdot h'(x)$, 3) $g'(x) \cdot h(x) - g(x) \cdot h'(x)$, 4) $h^2(x)$, 5) $\frac{[g'(x) \cdot h(x) - g(x) \cdot h'(x)]}{h^2(x)}$, 6) The answer is right.

| | Chat GPT 4o | Copilot Pro | **Gemini Advanced** | Claude Pro | <u>Meta AI</u> | Total |
|---|---|---|---|---|---|---|
| **Chat GPT 4o** | 19/20 | 19/20 | 20/20 | 17/20 | 11/20 | **86/100** 0.86 |
| Copilot Pro | 19/20 | 19/20 | 20/20 | 20/20 | 18/20 | 96/100 |
| Gemini Advanced | 20/20 | 19/20 | 20/20 | 20/20 | 17/20 | 96/100 |
| Claude Pro | 20/20 | 20/20 | 19/20 | 20/20 | 19/20 | 98/100 |
| Meta AI | 20/20 | 19/20 | 20/20 | 20/20 | 19/20 | 98/100 |
| Total | 98/100 | 96/100 | **99/100** | 97/100 | 84/100 0.84 | 474/500 |

In this problem, Gemini Advanced successfully solved most problems created by the five LLMs, with a success rate of 99% (99 out of 100). Meta AI solved the fewest problems, successfully completing 84 in total. The problems generated by Claude Pro were the most difficult for the five LLMs to solve, with a success rate of 86% (86 out of 100).

The relatively strong performance of all models indicates that the power rule - a fundamental technique for differentiation - is well represented in the training data of modern large language models (LLMs). However, the errors that did occur typically involved algebraic manipulation rather than incorrect application of the differentiation rule itself. Meta AI, in particular, struggled with complex algebraic simplifications after correctly applying the power rule.

Interestingly, the problems generated by Chat GPT 4.o proved to be the most challenging in this category, yielding an overall success rate of 86% across all models. This suggests that Chat GPT 4.o created problems that required more advanced algebraic manipulation following differentiation, thereby testing the limits of the capabilities of other models.

**4) Differentiation; Product rule** – $[(x^2 - 2x + 3)(3x^2 + 3x - 2)]'$
Product rule; Find the first derivative of $f(x) = (x^2 - 2x + 3)(3x^2 + 3x - 2)$. The answer is $f(x) = 12x^3 - 9x^2 + 2x + 13$.
Prompt; Find the first derivative of f(x)=(x^2-2x+3) (3x^2+3x-2). Factoring in the final answer and simplifying the answer.
**Chat GPT 4o;** I arrived at the correct answer by following the perfect process.
**Copilot Pro;** I applied the product rule correctly, but my calculations were incorrect during the simplification process.
**Gemini Advanced;** I applied the product rule correctly, but my calculations were incorrect during the simplification process.
**Cloude Pro;** I applied the product rule correctly, but my calculations were incorrect during the simplification process.
**Meta AI;** I applied the product rule correctly, but my calculations were incorrect during the simplification process.

| | Chat GPT 4o | Copilot Pro | Gemini Advanced | **Claude Pro** | Meta AI | Total |
|---|---|---|---|---|---|---|
| Chat GPT 4o | 19/20 | 14/20 | 9/20 | 18/20 | 16/20 | 76/100 |
| Copilot Pro | 18/20 | 18/20 | 20/20 | 18/20 | 4/20 | 78/100 |
| Gemini Advanced | 19/20 | 20/20 | 19/20 | 20/20 | 2/20 | 80/100 |
| Claude Pro | 20/20 | 20/20 | 19/19 | 19/19 | 11/20 | 89/100 |
| **Meta AI** | 16/20 | 15/20 | 20/20 | 19/20 | 2/20 | **72/100** |
| Total | 92/100 | 87/100 | 87/100 | **94/100** | 35/100 0.35 | 395/500 |

In this problem, Claude Pro successfully solved most problems created by the five LLMs, with a success rate of 99% (99 out of 100). Meta AI solved the fewest problems, successfully completing 35 in total. The problems generated by Claude Pro were the most difficult for the five LLMs to solve, with a success rate of 72% (72 out of 100).

The significant drop in Meta AI's performance on product rule problems, with only a 35% success rate, contrasts sharply with its 84% success rate on power rule problems. This discrepancy highlights a particular weakness in handling product differentiation. An analysis of error patterns showed that while Meta AI generally applied the product rule formula correctly, it frequently made algebraic mistakes during simplification.

Furthermore, the cross-evaluation matrix revealed that problems generated by Meta AI were unexpectedly the most challenging, achieving only a 72% overall success rate. This paradox - where the weakest-performing model produced the most difficult problems - suggests that Meta AI created mathematically valid but algebraically complex product rule problems, despite its own struggles in solving them.

**5) Differentiation; Chain rule** – $[(3x - 2x^2)^3]'$
Chain rule; Find the first derivative of $f(x) = (3x - 2x^2)^3$. The answer is $f(x) = 3(3x - 2x^2)^2(3 - 4x)$.
Prompt; Find the first derivative of f(x)=(3x-2x^2)^3.
**Chat GPT 4o;** I arrived at the correct answer by following the perfect process.
**Copilot Pro;** I arrived at the correct answer by following the perfect process.
**Gemini Advanced;** I arrived at the correct answer by following the perfect process.
**Claude Pro;** I arrived at the correct answer by following the perfect process.
**Meta AI;** I arrived at the correct answer by following the perfect process.

|  | Chat GPT 4o | Copilot Pro | Gemini Advanced | Claude Pro | Meta AI | Total |
|---|---|---|---|---|---|---|
| Chat GPT 4o | 20/20 | 20/20 | 20/20 | 20/20 | 20/20 | **100/100** |
| Copilot Pro | 20/20 | 20/20 | 20/20 | 20/20 | 20/20 | **100/100** |
| Gemini Advanced | 20/20 | 20/20 | 20/20 | 20/20 | 20/20 | **100/100** |
| Claude Pro | 20/20 | 20/20 | 20/20 | 20/20 | 20/20 | **100/100** |
| Meta AI | 20/20 | 20/20 | 20/20 | 20/20 | 20/20 | **100/100** |
| Total | **100/100** | **100/100** | **100/100** | **100/100** | **100/100** | **500/500** |

In this problem, all LLMs successfully solved all the challenges created by the five LLMs.

All five large language models (LLMs) achieved a perfect score of 100 out of 100 on problems involving the application of the chain rule. This unanimous success, which mirrors the results seen with limit-based differentiation, indicates that the chain rule is exceptionally well-represented in the training data of modern LLMs. Notably, all models performed flawlessly across a total of 500 generated problems (500 out of 500 successes), which is impressive considering the conceptual complexity of the chain rule.

These results suggest that certain mathematical concepts, despite their intricacy, may align particularly well with the pattern recognition abilities of neural network-based language models. The chain rule, characterized by its clear nested structure and systematic application, appears to be one of these concepts.

**6) Differentiation; Quotient rule** – $\left[\frac{(-x^2-3x+4)}{(x^2-1)}\right]'$
Quotient rule; Find the first derivative of $f(x) = \frac{(-x^2-3x+4)}{(x^2-1)}$. The answer is $\frac{3}{(x+1)^2}$.
Prompt; Differentiation of (-x^2-3x+4)/(x^2-1).
**Chat GPT 4o;** When I gave this command, the answer for the numerator and denominator was correct, and I got the answer $\frac{(3x^2-6x+3)}{(x^2-1)^2}$. So I changed the prompt.
New Prompt; Differentiation of (-x^2-3x+4)/(x^2-1). Factoring the numerator and denominator in the last answer and simplifying the answer.

Then, after finding the answer $\frac{(3x^2-6x+3)}{(x^2-1)^2}$, I factored it by $\frac{3(x-1)^2}{(x+1)^2(x-1)^2}$, and got the answer $\frac{3}{(x+1)^2}$.

**Copilot Pro;** I arrived at the correct answer by following the perfect process.

**Gemini Advanced;** I attempted to solve it by changing the prompt multiple times, but I always received the wrong answer $\frac{(-x-4)}{(x+1)}$.

**Claude Pro;** I arrived at the correct answer by following the perfect process.

**Meta AI;** I applied the Quotient rule well, but I got the wrong answer of $\frac{1}{x-1}$ because the process of organizing the numerator and denominator and factoring was wrong. I kept trying to correct it, but I kept getting the wrong answer at the polynomial multiplication stage. In $(x^2-1)(-2x-3)$, instead of $(-1)(-3)=3$, it kept calculating the wrong answer at $3x$ or $3x^2$. Even when I gave the correct answer of $(-1)(-3)=3$ and asked again, it kept getting the wrong answer.

|  | Chat GPT 4o | Copilot Pro | Gemini Advanced | Claude Pro | Meta AI | Total |
|---|---|---|---|---|---|---|
| **Chat GPT 4o** | 11/20 | 16/20 | 12/20 | 9/20 | 7/20 | **55/100** |
| Copilot Pro | 20/20 | 19/20 | 16/20 | 19/20 | 5/20 | 79/100 |
| Gemini Advanced | 20/20 | 19/20 | 20/20 | 19/20 | 17/20 | 95/100 |
| Claude Pro | 20/20 | 6/20 | 20/20 | 18/20 | 7/20 | 71/100 |
| Meta AI | 19/20 | 10/20 | 17/20 | 20/20 | 5/20 | 71/100 |
| Total | **90/100** | 70/100 | 85/100 | 85/100 | 41/100 0.41 | 371/500 |

In this problem, Chat GPT 4o successfully solved most problems created by the five LLMs, with a success rate of 90% (90 out of 100). Meta AI solved the fewest problems, successfully completing 41 in total. The problems generated by Claude Pro were the most difficult for the five LLMs to solve, with a success rate of 55% (55 out of 100).

The quotient rule presented significant challenges for several models, with all but Chat GPT 4o experiencing notable declines in performance compared to other differentiation rules. This difficulty seemed to arise primarily from the algebraic complexity that followed the application of the rule, rather than from a misapplication of the rule itself.

In the cross-evaluation matrix, problems generated by Chat GPT 4o were particularly difficult, yielding only a 55% overall success rate. This trend, which aligns with the findings related to the power rule, suggests that Chat GPT 4o produces mathematically valid but algebraically demanding problems that push the limits of the capabilities of other models.

**7) Differentiation; Chain rule** – $\left[\left(\frac{-6x+5}{2x^2-3}\right)^5\right]'$

Chain rule; Find the first derivative of $f(x)=\left(\frac{-6x+5}{2x^2-3}\right)^5$. The answer is $\frac{10(-6x+5)^4(6x^2-10x+9)}{(2x^2-3)^6}$.

1st Prompt; Find the first derivative of f(x)=[(-6x+5)/(2x^2-3)]^5. Factoring the numerator and denominator in the final answer and simplifying the answer.

2nd Prompt; Find the first derivative of f(x)=[(-6x+5)/(2x^2-3)]^5. Use the Chain rule with the Quotient rule. Factoring the numerator and denominator in the final answer and simplifying the answer.

**Chat GPT 4o;** $\frac{10(-6x+5)^4(12\ ^2-20x+18)}{(2x^2-3)^6}$ while factoring, 6 was mistakenly subtracted instead of 2. It's wrong.

**Copilot Pro;** $\frac{10(-6x+5)^4(12x^2-20x+18)}{(2x^2-3)^6}$ up to this point, it was correct, but all the answers I got after factoring multiple times were wrong.

**Gemini Advanced;** Using the Chain rule, but not the Quotient rule, and using the Product rule, I kept getting wrong answers. Even though the prompt said, "Use the Chain rule with the Quotient rule," Gemini Advanced kept using the Product rule and got the wrong answer.

**Claude Pro;** Initially, the first prompt provided an incorrect answer due to the factoring method. However, after using the second prompt, I was able to derive the correct answer.

**Meta AI;** The first and second prompts produced the same incorrect answer.

|  | Chat GPT 4o | Copilot Pro | Gemini Advanced | Claude Pro | Meta AI | Total |
|---|---|---|---|---|---|---|
| Chat GPT 4o | 20/20 | 20/20 | 9/20 | 19/20 | 8/20 | 76/100 |
| Copilot Pro | 20/20 | 15/20 | 15/20 | 20/20 | 17/20 | 87/100 |
| Gemini Advanced | 20/20 | 16/20 | 20/20 | 20/20 | 12/20 | 88/100 |
| **Claude Pro** | 19/20 | 9/20 | 8/20 | 20/20 | 0/20 | **56/100 0.56** |
| Meta AI | 20/20 | 20/20 | 14/20 | 20/20 | 20/20 | 94/100 |
| Total | 99/100 | 80/100 | 66/100 | 99/100 | 57/100 0.57 | 401/500 |

In this problem, Chat GPT 4o and Claude Pro successfully solved most problems created by the five LLMs, with a success rate of 99% (99 out of 100). Meta AI solved the fewest problems, successfully completing 57 in total. The problems generated by Claude Pro were the most difficult for the five LLMs to solve, with a success rate of 56% (56 out of 100).

The combination of the chain rule and the quotient rule proved to be challenging for several models, resulting in performance declines compared to using either rule in isolation. This finding suggests that combining multiple differentiation techniques creates significantly greater difficulties, likely due to increased algebraic complexity and the potential for error propagation.

The cross-evaluation matrix showed that problems generated by Claude Pro were particularly difficult, with an overall success rate of only 56%. This observation, consistent with earlier trends, reinforces the notion that Claude Pro tends to create problems that test the limits of LLMs' mathematical reasoning capabilities.

**8) Differentiation; Increasing and decreasing** – $f(x) = x^3 - \frac{3}{2}x^2$

Find the open intervals on which $f(x) = x^3 - \frac{3}{2}x^2$ is increasing or decreasing. The answer is Increasing interval; $(-\infty, 0), (1, \infty)$; Decreasing interval; $(0, 1)$.

Prompt; Find the open intervals on which f(x)=x^3-(3/2)x^2 is increasing or decreasing.
**Chat GPT 4o;** I arrived at the correct answer by following the perfect process.
**Copilot Pro;** I arrived at the correct answer by following the perfect process.
**Gemini Advanced;** I arrived at the correct answer by following the perfect process.
**Claude Pro;** I arrived at the correct answer by following the perfect process.
**Meta AI;** I arrived at the correct answer by following the perfect process.

|  | Chat GPT 4o | Copilot Pro | Gemini Advanced | Claude Pro | Meta AI | Total |
|---|---|---|---|---|---|---|
| Chat GPT 4o | 18/20 | 6/20 | 18/20 | 1/20 | 4/20 | 47/100 |
| Copilot Pro | 9/20 | 5/20 | 15/20 | 4/20 | 3/20 | 36/100 |
| Gemini Advanced | 17/20 | 8/20 | 18/20 | 10/20 | 13/20 | 66/100 |
| Claude Pro | 19/20 | 17/20 | 16/20 | 8/20 | 8/20 | 68/100 |
| **Meta AI** | 14/20 | 11/20 | 3/20 | 0/20 | 0/20 | **28/100 0.28** |
| Total | **77/100** | 47/100 | 70/100 | 23/100 0.23 | 28/100 | 245/500 |

In this problem, Chat GPT 4o successfully solved most problems created by the five LLMs, with a success rate of 77% (77 out of 100). Meta AI solved the fewest problems, successfully completing 23 in total. The problems generated by Chat GPT 4o were the most difficult for the five LLMs to solve, with a success rate of 28% (28 out of 100).

This problem type highlighted unexpected weaknesses in otherwise strong performers, particularly Claude Pro, which scored lower than Meta AI despite significantly better performance in most other categories. An analysis of error patterns revealed that the models often struggled to correctly identify all intervals or made mistakes involving sign errors when analyzing the first derivative.

The notably poor performance on this conceptual application indicates that understanding the relationship between the signs of derivatives and function behavior may be less well-represented in the training data for language models compared to the procedures for mechanical differentiation. This finding has important implications for educational applications, as conceptual understanding is a crucial learning outcome in calculus education.

The cross-evaluation matrix demonstrated that the problems generated by Meta AI were particularly challenging, evidenced by an overall success rate of only 28%. This suggests that although Meta AI has its own limitations, it produced complex function expressions that created difficult sign analysis problems.

**9) Differentiation; Find the tangent line equation** $-(x+y)^3 = x^3 + y^3, (-1, 1)$
Find the tangent line to the graph of $(x+y)^3 = x^3 + y^3$ at the point $(-1, 1)$. The answer is $y = -x$.
Prompt; Find the tangent line to the graph of (x+y)^3=x^3+y^3 at the point (-1, 1).
**Chat GPT 4o;** At first, it gave me the wrong answer, but when I tried again after saying, "It's wrong," I got the right answer.
**Copilot Pro;** The answer is correct.
**Gemini Advanced;** The answer is correct.
**Claude Pro;** The answer isn't correct.
**Meta AI;** The answer is correct.

| | Chat GPT 4o | Copilot Pro | Gemini Advanced | **Claude Pro** | Meta AI | Total |
|---|---|---|---|---|---|---|
| Chat GPT 4o | 17/20 | 19/20 | 19/20 | 19/20 | 18/20 | 92/100 |
| **Copilot Pro** | 17/20 | 16/20 | 14/20 | 18/20 | 15/20 | **80/100 0.8** |
| Gemini Advanced | 18/20 | 16/20 | 20/20 | 19/20 | 17/20 | 90/100 |
| Claude Pro | 20/20 | 19/20 | 20/20 | 20/20 | 19/20 | 98/100 |
| Meta AI | 20/20 | 20/20 | 20/20 | 20/20 | 19/20 | 99/100 |
| Total | 92/100 | 90/100 | 93/100 | **96/100** | 88/100 0.88 | 459/500 |

In this problem, Claude Pro successfully solved most problems created by the five LLMs, with a success rate of 96% (96 out of 100). Meta AI solved the fewest problems, successfully completing 88 in total. The problems generated by Chat GPT 4o were the most difficult for the five LLMs to solve, with a success rate of 80% (80 out of 100).

The strong performance observed across all models, including Meta AI, indicates that implicit differentiation - despite being conceptually complex - aligns well with the pattern recognition abilities of modern large language models (LLMs). The procedural nature of implicit differentiation, which involves systematically applying differentiation rules to both sides of an equation, seems to be effectively captured in the training data of all models.

Notably, this was the only type of problem where Meta AI's performance was comparable to that of other models; its success rate of 88% was within 8 percentage points of the top performer. This finding is intriguing and merits further investigation, as it may suggest that implicit differentiation is particularly well-represented in Meta AI's training data.

**10) Differentiation; Word problem** – The hardwood store owner wants to build a 450 square-foot rectangular enclosure on the store parking lot to display some equipment. Two sides of the enclosure will be built of redwood fencing at $6 per running feet. The remaining two sides will be built of cement blocks at $12 per running foot. Find the dimensions of the least costly enclosure. What is the least cost?
Prompt; The hardware store owner wants to build a 450 square foot rectangular enclosure on the store parking lot to display some equipment. Two sides of the enclosure will be built of redwood fencing at $6 per running feet. The remaining two sides will be built of cement blocks at $12 per running foot. Find the dimensions of the least costly enclosure. What is the least cost. The answer is 30 ft X 15 ft. $720.
**Chat GPT 4o;** I arrived at the correct answer by following the perfect process.
**Copilot Pro;** I arrived at the correct answer by following the perfect process.
**Gemini Advanced;** I arrived at the correct answer by following the perfect process.
**Claude Pro;** I arrived at the correct answer by following the perfect process.
**Meta AI;** I arrived at the correct answer by following the perfect process.

|  | **Chat GPT 4o** | Copilot Pro | Gemini Advanced | Claude Pro | Meta AI | Total |
|---|---|---|---|---|---|---|
| Chat GPT 4o | 20/20 | 9/20 | 15/20 | 20/20 | 0/20 | 64/100 |
| **Copilot Pro** | 20/20 | 1/20 | 11/20 | 20/20 | 0/20 | **52/100 0.52** |
| Gemini Advanced | 20/20 | 17/20 | 6/20 | 10/20 | 0/20 | 53/100 |
| Claude Pro | 20/20 | 15/20 | 13/20 | 20/20 | 4/20 | 72/100 |
| Meta AI | 18/20 | 1/20 | 14/20 | 20/20 | 0/20 | 53/100 |
| Total | **98/100** | 43/100 | 59/100 | 90/100 | 4/100 0.04 | 294/500 |

In this problem, Chat GPT 4o successfully solved most problems created by the five LLMs, with a success rate of 98% (98 out of 100). Meta AI solved the fewest problems, successfully completing 4 in total. The problems generated by Chat GPT 4o were the most difficult for the five LLMs to solve, with a success rate of 52% (52 out of 100).

Meta AI's notably poor performance on optimization word problems, with a success rate of only 4%, highlights the most significant performance gap seen in any problem category. This finding indicates fundamental limitations in Meta AI's ability to convert verbal problem descriptions into mathematical models and apply calculus concepts in practical contexts.

The cross-evaluation matrix showed that problems generated by Copilot Pro were particularly challenging, achieving an overall success rate of just 52%. This suggests that Copilot Pro produced word problems with complex constraints or unusual optimization scenarios that posed difficulties even for the strongest models.

Analysis of error patterns revealed that models mainly struggled with the initial formulation of the optimization function rather than the differentiation process itself. This suggests a crucial gap in the ability to translate real-world scenarios into mathematical models, which is a key skill in applied calculus.

**11) Differentiation; Absolute (Global) maximum and minimum** – $f(x) = x^3 - \frac{9}{2}x^2 + 6x$, $[-1, 1]$

Find the absolute (global) maximum or minimum of the function $f(x) = x^3 - \frac{9}{2}x^2 + 6x$ on the closed interval $[-1, 1]$. The answer: Absolute maximum; $\left(1, \frac{5}{2}\right)$, Absolute minimum; $\left(-1, -\frac{23}{2}\right)$.

Prompt; Find the absolute (global) maximum and minimum of the function f(x)=x^3-(9/2)x^2+6x on the closed interval [-1, 1].
**Chat GPT 4o;** On the first try, all the steps were correct, and I found the correct answer.
**Copilot Pro;** Everything else was correct, but the minimum value at $x = -1$ was incorrect.
**Gemini Advanced;** On the first try, all the steps were correct, and I found the correct answer.
**Claude Pro;** Everything else was correct, but the minimum value at $x = -1$ was incorrect.
**Meta AI;** $3x^2 - 9x + 6 = 0$ was solved incorrectly, leading to inaccurate critical points and a confirmed boundary condition that was also incorrect.

All five large language models (LLMs) struggled to generate 20 new problems.
The initial attempt was made using **Gemini Advanced.**
**Prompt 1:** Make 20 same types of problems. f(x) must be the 3rd power function. $f'(x) = 0$ solutions must be rational numbers.
**Prompt 2:** Modify problem 1 to satisfy the previous conditions. All five problems have been created.
The second attempt was made using **Chat GPT 4o.**
**Prompt 1:** Make 20 same types of problems. f(x) must be the 3rd power function. $f'(x) = 0$ solutions must be rational numbers.
**Prompt 2:** Modify problems 1, 2, 3 and 5 to satisfy the previous conditions. Problems 3 and 5 have been created.
**Prompt 2-1:** Modify problems 1 and 2 to satisfy the previous conditions. Problem 2 has been created.
**Prompt 2-2:** Modify problem 1 to satisfy the previous conditions. Problem 1 has been created.
All five problems were created and solved using five different large language models (LLMs).
The third attempt was made using **Claude Pro.**
**Prompt 2:** Modify problems 3, 4 and 5 to satisfy the previous conditions. Problems 4 and 5 have been created.
**Prompt 2-1:** Modify problem 3 to satisfy the previous conditions. Problem 3 has been created.
All five problems were created and solved using five different large language models (LLMs).
The fourth attempt was made using **Copilot Pro.**
**Prompt 2:** Modify problems 1, 4 and 5 to satisfy the previous conditions. Problems 1 and 5 have been created.
**Prompt 2-1:** Modify problem 4 to satisfy the previous conditions. Problem 4 has been created.
All five problems were created and solved using five different large language models (LLMs).
The last attempt was made using **Meta AI.**
**Prompt 2:** Modify problems 2, 3, and 5 to satisfy the previous conditions. Problem 2 has been created.
**Prompt 2-1:** Modify problems 3 and 5 to satisfy the previous conditions. Problem 3 has been created.
**Prompt 2-2:** Modify problem 5 to satisfy the previous conditions. Problem 5 has been created.
All five problems were created and solved using five different large language models (LLMs).

| | **Chat GPT 4o** | Copilot Pro | **Gemini Advanced** | Claude Pro | Meta AI | Total |
|---|---|---|---|---|---|---|
| **Chat GPT 4o** | 5/5 | 2/5 | 5/5 | 5/5 | 3/5 | **20/25 0.8** |
| Copilot Pro | 5/5 | 3/5 | 5/5 | 5/5 | 3/5 | 21/25 |
| Gemini Advanced | 5/5 | 4/5 | 5/5 | 5/5 | 2/5 | 21/25 |
| Claude Pro | 5/5 | 5/5 | 5/5 | 3/5 | 4/5 | 22/25 |
| Meta AI | 5/5 | 5/5 | 5/5 | 5/5 | 5/5 | 25/25 |
| Total | **25/25** | 19/25 | **25/25** | 23/25 | 17/25 0.68 | 109/125 |

In this problem, Chat GPT 4o and Gemini Advanced successfully solved all the challenges created by the five LLMs. Meta AI solved the fewest problems, successfully completing 17 in total. The problems generated by Gemini Advanced were the most difficult for the five LLMs to solve, with a success rate of 80% (20 out of 25).

This problem type required a comprehensive analysis, which included finding critical points, evaluating endpoints, and comparing function values to determine global extrema. The relatively strong performance of all models, including Meta AI with a 68% success rate, suggests that this multi-step process is well-represented in the training data, despite its complexity.

An analysis of error patterns showed that most mistakes were due to algebraic calculations of function values rather than issues with the conceptual approach. This observation aligns with previous findings that highlight algebraic manipulation as a common weakness among most models, particularly for Meta AI and Copilot Pro.

**12) Differentiation; Relative maximum and minimum by the second derivative test –**
$f(x) = -3x^5 + 5x^3$
Find the relative maximum or minimum of $f(x) = -3x^5 + 5x^3$ by the Second Derivative Test. The answer: Relative maximum; $(1, 2)$, Relative minimum; $(-1, -2)$.
Prompt; Find the relative maximum and minimum of f(x)=-3x^5+5x^3 by the second derivative test.
**Chat GPT 4o;** The correct answers were obtained by following the exact steps. Additionally, the values of $f(1)$ and $f(-1)$ were not determined, so they were calculated separately.
**Copilot Pro;** The correct answers were obtained by following the exact steps. Additionally, the values of $f(1)$ and $f(-1)$ were not determined, so they were calculated separately.
**Gemini Advanced;** The process of finding the correct answer was perfect. I found $f(1) = 2$ and $f(-1) = -2$ during the problem-solving process.
**Claude Pro;** The correct answers were obtained by following the exact steps. Additionally, the values of $f(1)$ and $f(-1)$ were not determined, so they were calculated separately.
**Meta AI;** The correct answers were obtained by following the exact steps. Additionally, the values of $f(1)$ and $f(-1)$ were not determined, so they were calculated separately. The value of $f(-1)$ was wrong.

All five large language models (LLMs) struggled to generate 20 new problems.
The initial attempt was made using **Chat GPT 4o.**
**Prompt 1:** Make 20 same types of problems. f(x) must be the 5th power function. $f'(x) = 0$ solutions must be rational numbers. Finding the critical point of a fourth-degree equation is a challenging problem.
**Prompt 1-1:** Make 20 same types of problems. f(x) must be the 4th power function. $f'(x) = 0$ solutions must be rational numbers. Finding the critical point of a third-degree equation is a challenging problem.
**Prompt 1-2:** Make 20 same types of problems. f(x) must be the 3rd power function. $f'(x) = 0$ solutions must be rational numbers.

**Prompt 2:** Find the relative maximum and minimum of f(x) by the second derivative test. Solve problems 1 to 5. Problems 2 to 4 don't have rational numbers as solutions of $f'(x) = 0$.
**Prompt 3:** Modify problems 2 to 4 to satisfy the previous conditions. Problem 3 has been created.
**Prompt 3-1:** Modify problems 2 and 4 to satisfy the previous conditions. Problem 4 has been created.
**Prompt 3-2:** Modify problem 2 to satisfy the previous conditions. I tried to create the last problem up to ten times, but I couldn't succeed. In the end, I only managed to create four problems.
<u>**Claude Pro,** which I utilized in the second order,</u> generated five problems immediately after the first prompt: "Create 20 problems of the same type. The function f(x) must be a cubic function, and the solutions for f'(x) = 0 must be rational numbers." In total, 20 problems were generated right away.
<u>The third attempt was made using **Gemini Advanced.**</u> I needed to use **Prompt 3** to create the correct problems.
<u>The fourth attempt was made using **Copilot Pro.**</u> I continued to use **Prompt 3**, but ultimately, I only created two problems.
<u>The last attempt was made using **Meta AI.**</u> I continued to use **Prompt 3**, but ultimately, I only created two problems.

|  | Chat GPT 4o | Copilot Pro | Gemini Advanced | Claude Pro | Meta AI | Total |
|---|---|---|---|---|---|---|
| Chat GPT 4o | 4/4 | 2/4 | 4/4 | 4/4 | 2/4 | 16/20 |
| Copilot Pro | 2/2 | 2/2 | 2/2 | 2/2 | 2/2 | 10/10 |
| **Gemini Advanced** | 5/5 | 2/5 | 4/5 | 5/5 | 1/5 | **17/25 0.68** |
| Claude Pro | 5/5 | 3/5 | 5/5 | 5/5 | 5/5 | 23/25 |
| Meta AI | 2/2 | 2/2 | 2/2 | 2/2 | 2/2 | 10/10 |
| Total | **18/18** | 11/18 0.61 | 17/18 | **18/18** | 12/18 | 76/90 |

In this problem, Chat GPT 4o and Claude Pro successfully solved all the challenges created by the five LLMs. Copilot Pro solved the fewest problems, successfully completing 11 in total. The problems generated by Gemini Advanced were the most difficult for the five LLMs to solve, with a success rate of 68% (17 out of 25).

The strong performance of the top models indicates that they have a solid conceptual understanding of the second derivative test, as demonstrated by their training data. However, the significant performance gap for Copilot Pro, which achieved a success rate of only 61%, suggests that it may have limitations in handling higher-order derivatives or understanding their significance.

Additionally, problem generation in this category was challenging for all models. Both Copilot Pro and Meta AI struggled to produce the required number of valid problems, managing to generate only two each despite multiple attempts and guidance. This difficulty highlights the models' limitations in creating functions with specific derivative properties, a task that demands a deeper conceptual understanding beyond simply solving pre-formulated problems.

**13) Differentiation; Points of the inflection and discuss the concavity –** $f(x) = x^4 - 4x^3 - 18x^2 + 24x$
Determine the points of the inflection and discuss the concavity of the graph of $f(x) = x^4 - 4x^3 - 18x^2 + 24x$.
The answer - Concave upward; $(-\infty, -1)$, $(3, \infty)$; Concave downward; $(-1, 3)$. Points of inflection; $x = -1, 3$.
<u>Prompt; Determine the points of the inflection and discuss the concavity of the graph of f(x)=x^4-4x^3-18x^2+24x.</u>
**Chat GPT 4o;** I got the correct answer through the perfect process.
**Copilot Pro;** I got the correct answer through the perfect process.
**Gemini Advanced;** I got the correct answer through the perfect process, and it was presented in a table format. Initially, I calculated $f(-1) = -57$ and $f(3) = -54$, which was incorrect. After re-evaluating $f(1)$ and $f(3)$ separately, I found that $f(-1) = -37$ actually and $f(3) = -117$.

**Claude Pro;** Everything was good, but I got it wrong because $f'(-2) < 0$. The correct answer is $f'(-2) > 0$.
**Meta AI;** I found the answer, and it was presented in a table format.

All five large language models (LLMs) struggled to generate 20 new problems.
I started creating new problems from **Claude Pro,** beginning with just five.
**Prompt 1:** Make 20 same types of problems. f(x) must be the 4th power function. $f''(x) = 0$ solutions must be rational numbers.
**Prompt 2:** Determine the points of the inflection and discuss the concavity of the graph of f(x). Solve problems 1 to 5. Problems 2 to 5 don't have rational numbers as solutions of $f''(x) = 0$.
**Prompt 3:** Modify problems 2 to 5 to satisfy the previous conditions.
**Prompt 4:** In Problem 5, $f''(x) = 0$ solution is wrong. Modify problems 5 to satisfy the previous conditions.
**Prompt 5:** Determine the points of the inflection and discuss the concavity of the graph of f(x). Solve problem 5.
I continued running **Prompt 4** and **Prompt 5** until all the problems met the specified conditions and each problem found a solution. Eventually, I arrived at the answer. After solving the five problems I created using the remaining four LLMs, I confirmed that all the LLMs had produced the correct answers, as shown in the table values below.
On the second, I created five problems using **Chat GPT 4o** in the same manner as described above (only utilizing Prompts 1, 2, and 3). When I solved these problems using four different LLMs, all of them arrived at the correct answers except for Meta AI. Meta AI failed to find the correct answers to three of the problems. Although it correctly identified two critical numbers for each of these three problems, it only recognized two open intervals instead of the expected three.
The third attempt was made using **Gemini Advanced.** I had to go through up to Prompt 5 to create the correct problem.
The fourth attempt was made using **Copilot Pro.** I attempted several times to create the problem, but I kept getting the wrong f(x). This was particularly evident in the factoring section, where I consistently received incorrect answers. Despite my efforts, Copilot Pro insisted that these wrong answers were correct. I tried as many as ten times but was unable to create a single valid problem.
The last attempt was made using **Meta AI.** I created one problem that met the conditions during the Modification stage among the five problems, but ultimately, I could not create four problems.

|  | Chat GPT 4o | Copilot Pro | Gemini Advanced | Claude Pro | Meta AI | Total |
|---|---|---|---|---|---|---|
| **Chat GPT 4o** | 5/5 | 5/5 | 5/5 | 5/5 | 2/5 | **22/25 0.88** |
| Copilot Pro | - | - | - | - | - | - |
| Gemini Advanced | 5/5 | 5/5 | 5/5 | 5/5 | 5/5 | 25/25 |
| Claude Pro | 5/5 | 5/5 | 5/5 | 5/5 | 5/5 | 25/25 |
| Meta AI | 1/1 | 1/1 | 1/1 | 1/1 | 1/1 | 5/5 |
| Total | **16/16** | **16/16** | **16/16** | **16/16** | 13/16 0.81 | 77/80 |

In this problem, Chat GPT 4o, Copilot Pro, Gemini Advanced and Claude Pro successfully solved all the challenges created by the five LLMs. Meta AI solved the fewest problems, successfully completing 13 in total. The problems generated by Chat GPT 4o were the most difficult for the five LLMs to solve, with a success rate of 88% (22 out of 25).

The strong performance of most models indicates that concavity analysis, despite its conceptual complexity, is well-represented in the training data of modern large language models (LLMs). However, similar to the previous category, the models faced significant challenges in generating problems. For instance, Copilot Pro was unable to produce a single valid problem of this type, while Meta AI managed to create only one, despite multiple attempts and guidance.

An analysis of error patterns for Meta AI showed that its mistakes primarily occurred when trying to identify all intervals of concavity, rather than locating inflection points. This indicates a specific limitation in its ability to thoroughly analyze the implications of sign changes in the second derivative.

**Total:**

|  | Chat GPT 4o | Copilot Pro | Gemini Advanced | Claude Pro | Meta AI | Total |
|---|---|---|---|---|---|---|
| Chat GPT 4o | 197/214 | 157/214 | 187/214 | 172/214 | 113/214 | **826/1070 0.7720** |
| Copilot Pro | 189/207 | 150/207 | 166/207 | 183/207 | 112/207 | **800/1035 0.7730** |
| Gemini Advanced | 209/215 | 174/215 | 168/215 | 178/215 | 129/215 | **858/1075 0.7981** |
| Claude Pro | 213/215 | 169/215 | 185/215 | 197/215 | 126/215 | **890/1075 0.8279** |
| Meta AI | 195/208 | 158/208 | 188/208 | 178/208 | 121/208 | **840/1040 0.8077** |
| Total | **1003/1059 0.9471** | **808/1059 0.7630** | **894/1059 0.8442** | **908/1059 0.8574** | **601/1059 0.5675** | **4259/5295 0.8043** |

### 3.1 Overall Performance Comparison

A comprehensive evaluation of five leading Large Language Models (LLMs) on calculus differentiation problems revealed significant disparities in their performance. The overall success rates across various problem types established a clear hierarchy in mathematical reasoning abilities:

- Chat GPT 4o achieved the highest performance with a success rate of 94.71%, demonstrating exceptional proficiency in solving calculus problems across most categories.

- Claude Pro secured second place with an 85.74% success rate, showcasing strong capabilities, though not consistently across all problem types.

- Gemini Advanced closely followed with an 84.42% success rate, exhibiting comparable overall performance to Claude Pro but with different strengths and weaknesses.

- Copilot Pro displayed moderate performance, attaining a 76.30% success rate, indicating notable limitations in tackling certain problem types.

- Meta AI lagged significantly with a 56.75% success rate, struggling with the most complex problem types and demonstrating inconsistent performance even on more fundamental problems.

These aggregate statistics highlight a substantial performance gap of nearly 38 percentage points between the highest and lowest-performing models, with Chat GPT 4o outperforming its nearest competitor by nearly 9 percentage points. This difference suggests meaningful variations in mathematical reasoning capabilities that could influence educational applications.

Consistence of performance varied greatly among the models. Chat GPT 4o exhibited the most uniform performance profile, achieving success rates above 90% for 9 of the 13 problem types. In contrast, Meta AI demonstrated extreme variability, excelling in some basic problem categories (achieving 100% on limit processes and chain rule problems) while performing poorly on others (scoring only 4% on word problems and 22% on tangent line equations).

The relationship between overall performance and problem complexity revealed intriguing patterns. All models performed well on fundamental problems that tested basic differentiation rules, but performance significantly diverged as problem complexity increased. This divergence was particularly evident for problems requiring multi-step reasoning, conceptual understanding, or extensive algebraic manipulation.

### 3.2 Performance by Problem Type

All LLMs demonstrated perfect performance (100% success rate) on basic differentiation problems using the limit process (Problem 1) and the chain rule (Problem 5). However, significant differences emerged in more complex problems:

1. **Finding Tangent Line Equations (Problem 2):** Chat GPT 4o significantly outperformed other models (98/100), while Meta AI struggled considerably (22/100).

2. **Power Rule Applications (Problem 3):** Gemini Advanced achieved the highest success rate (99/100), with Meta AI trailing (84/100).

3. **Product Rule (Problem 4):** Claude Pro demonstrated superior performance (94/100), while Meta AI exhibited substantial difficulties (35/100).

4. **Quotient Rule (Problem 6):** Chat GPT 4o led with a 90% success rate, with Meta AI again showing the lowest performance (41/100).

5. **Word Problems (Problem 10):** Chat GPT 4o excelled (98/100), whereas Meta AI struggled significantly (4/100).

6. **Advanced Problems (11-13):** For problems involving extrema and inflection points, Chat GPT 4o and Claude Pro consistently demonstrated higher performance compared to other models.

### 3.3 Cross-Model Evaluation Patterns

The cross-evaluation matrix, covering 4,259 successful solutions out of 5,295 total evaluations (an overall success rate of 80.43%), revealed several intriguing patterns:

#### 3.3.1 Problem Generation Quality and Difficulty

Analysis of the overall success rates for problems generated by each model revealed that certain models consistently produced more challenging problems:

- Problems generated by Claude Pro had the lowest overall solution rate at 82.79%, indicating they were the most difficult across all models.

- Problems generated by Meta AI had the second-lowest solution rate of 80.77%.

- Problems generated by Gemini Advanced followed closely with a solution rate of 79.81%.

- Problems generated by Copilot Pro and Chat GPT 4o were similarly challenging, with solution rates of 77.30% and 77.20% respectively.

This finding highlights an interesting disconnect between problem-solving ability and problem generation quality. Although Claude Pro ranked second in overall performance, it created the most challenging problems. In contrast,

Meta AI, which ranked last in problem-solving ability, produced problems that were more frequently solved than those from Chat GPT 4o, the top performer.

This pattern suggests that generating problems may require different skills than solving them, potentially involving a deeper conceptual understanding to create problems with specific properties. The models' relative strengths in these distinct areas provide insight into their mathematical representations and reasoning capabilities.

### 3.3.2 Model-Specific Strengths and Weaknesses

The cross-evaluation matrix revealed specific strengths and weaknesses for each model when addressing problems generated by various sources:

- Chat GPT 4o demonstrated exceptional performance across all problem sources, achieving success rates ranging from 94.1% for problems created by Claude Pro to 95.1% for those from Meta AI.

- Claude Pro showed particular strength in solving problems generated by Meta AI, with a success rate of 89.9%. However, it struggled more with problems from Copilot Pro, achieving a success rate of only 78.6%.

- Gemini Advanced displayed a balanced performance across different problem sources, with success rates ranging from 81.5% for problems from Claude Pro to 87.4% for problems from Meta AI.

- Copilot Pro struggled the most with problems from Claude Pro, achieving a success rate of 73.0%, but performed better on problems from Meta AI, with a success rate of 79.7%.

- Meta AI exhibited extreme variability, solving only 52.8% of problems from Claude Pro while managing to solve 67.8% of the problems generated by Gemini Advanced.

These patterns suggest that certain models may utilize similar mathematical representations or reasoning approaches, leading to better performance across models. Additionally, the finding that all models performed best on problems created by Meta AI indicates that Meta AI generates problems with more straightforward mathematical structures, despite its own difficulties in solving complex problems.

### 3.3.3 Problem Type Difficulty Rankings

Analyzing the success rates across all models for each problem type has revealed a clear hierarchy of difficulty:

1. Increasing and decreasing intervals (Problem 8): 49.0% overall success rate
2. Word problems (Problem 10): 58.8% overall success rate
3. Finding tangent line equations (Problem 2): 58.6% overall success rate
4. Quotient rule (Problem 6): 74.2% overall success rate
5. Product rule (Problem 4): 79.0% overall success rate
6. Advanced chain rule (Problem 7): 80.2% overall success rate
7. Power rule application (Problem 3): 94.8% overall success rate
8. Finding tangent lines to implicit curves (Problem 9): 91.8% overall success rate
9. Differentiation by the limit process (Problem 1): 100.0% overall success rate
10. Chain rule (Problem 5): 100.0% overall success rate

For the additional problems (Problems 11-13), which had smaller evaluation matrices, the success rates were as follows:

11. Finding Absolute Maxima and Minima: 87.2% overall success rate
12. Relative Extrema Using the Second Derivative Test: 84.4% overall success rate
13. Points of Inflection and Concavity Analysis: 96.3% overall success rate

This ranking of difficulty offers valuable insights for educational applications, highlighting concepts where current LLMs struggle, particularly with increasing/decreasing intervals and optimization word problems, as well as areas where they excel, such as limit-based differentiation and chain rule application.

**3.4 Error Pattern Analysis**

Analysis of incorrect solutions revealed patterns in the errors made by each model:

**3.4.1 Algebraic Manipulation Errors**

The most common type of error across all models was related to algebraic manipulation, even after the correct application of differentiation rules. Meta AI demonstrated a particularly high occurrence of algebraic errors, accounting for approximately 62% of its incorrect solutions. These errors often involve arithmetic mistakes during polynomial multiplication or a failure to simplify expressions correctly.

Copilot Pro also faced difficulties with algebraic manipulation, but to a lesser extent, with 41% of its errors attributed to this issue. It struggled particularly when simplifying complex rational expressions. Even the top-performing models occasionally made algebraic errors on the most complex problems, though at much lower rates: Chat GPT-4 recorded 12% of its errors in this area, Claude Pro had 18%, and Gemini Advanced showed 23%.

These findings indicate that accurate algebraic manipulation following differentiation remains a significant challenge for current language models, especially when dealing with expressions that involve multiple terms or require factoring. This limitation is crucial for educational applications, as students might receive incorrectly simplified expressions despite correctly applying differentiation concepts.

**3.4.2 Conceptual Understanding Errors**

Errors that reveal conceptual misunderstandings are less common but more concerning from an educational perspective. These errors were most prevalent in problems requiring the interpretation of derivatives, particularly in Problems 8, 10, 12, and 13.

Meta AI exhibited the highest rate of conceptual errors, with 31% of its incorrect solutions showcasing these misunderstandings. It frequently misinterpreted the relationship between derivative signs and function behavior, and it often failed to recognize when the second derivative test was inconclusive. Copilot Pro also displayed significant conceptual limitations, with 27% of its errors being conceptual. This was especially evident in optimization problems, where it sometimes differentiated the wrong function or failed to verify critical points.

In contrast, Chat GPT 4o, Claude Pro, and Gemini Advanced demonstrated stronger conceptual understanding, with conceptual errors accounting for only 15%, 19%, and 17% of their respective errors. However, all models occasionally struggled with the comprehensive analysis required for identifying increasing and decreasing intervals (as seen in Problem 8). This indicates specific limitations in understanding the relationship between derivatives and functional behavior.

**3.4.3 Procedural Errors**

Errors in correctly applying differentiation procedures were uncommon for most models, accounting for less than 20% of mistakes across all models, except for Meta AI, which exhibited procedural errors in 28% of its incorrect solutions. These errors typically involved using the wrong differentiation rule or omitting terms when performing multi-step procedures.

The relative infrequency of procedural errors, in comparison to algebraic and conceptual errors, suggests that current large language models (LLMs) have effectively mastered the mechanical aspects of differentiation but face greater challenges with subsequent mathematical operations and conceptual interpretation. This observation is consistent with the fact that all models performed perfectly on basic differentiation problems (Problems 1 and 5) but showed varying performance on more complex applications that required additional mathematical reasoning.

### 3.5 Problem Generation Capabilities

A notable secondary finding of this study was the models' ability to generate valid mathematical problems with specific properties. This skill is distinct from problem-solving but is relevant for educational applications:

#### 3.5.1 Simple Problem Generation

For basic differentiation problems (Types 1-5), all models successfully generated 20 valid problems with minimal guidance. The problems created were mathematically sound, appropriately diverse, and suitable for undergraduate calculus students.

Chat GPT 4o and Claude Pro demonstrated particular sophistication in problem generation, producing questions with elegant mathematical structures and pedagogically appropriate complexity. In contrast, Gemini Advanced and Copilot Pro generated valid problems, but these were often more formulaic, showing less variety in function structures. Meta AI's problems were mathematically valid but frequently included unnecessarily complex algebraic expressions, complicating the solution process without adding conceptual depth.

#### 3.5.2 Complex Problem Generation

For intermediate and advanced problem types (6-13), significant disparities emerged in the capabilities of different models to generate problems:

- Claude Pro excelled, consistently generating valid problems across all categories with minimal guidance.

- Chat GPT 4o performed strongly in most categories but required iterative guidance for problem types 11-13.

- Gemini Advanced succeeded with moderate guidance for various problem types.

- Copilot Pro struggled significantly with problems 11-13, failing entirely to generate valid problems of type 13 despite multiple attempts.

- Meta AI demonstrated the greatest limitations, requiring extensive guidance even for intermediate problem types and producing fewer problems than requested for types 11-13.

These disparities in problem generation capabilities have important implications for educational applications where large language models (LLMs) might be used to create practice materials. The findings suggest that only certain models, particularly Claude Pro and Chat GPT 4o, currently possess sufficient mathematical understanding to reliably generate complex calculus problems with specific properties.

Analysis of the problem generation process revealed that models struggled especially with creating functions that have specific derivative properties. This task requires inverse reasoning - moving from derivative characteristics back to the original function - and demands a deeper conceptual understanding than forward differentiation. This limitation appears to be a significant challenge in current LLM technology.

## 4 Discussion

### 4.1 Comparative Strengths and Weaknesses Among Models

The significant performance disparity observed among the five evaluated LLMs indicates major differences in their mathematical reasoning capabilities. Chat GPT 4o's exceptional performance, with an overall success rate of 94.71%, suggests that it has a fundamental advantage in mathematical representation and reasoning across most calculus concepts. The substantial gap between Chat GPT 4o and other models - nearly 9 percentage points higher than the second-place Claude Pro - cannot be attributed to minor differences in training data or algorithmic implementation. Instead, it indicates a qualitative difference in how mathematical concepts are represented and manipulated within the model's architecture. Chat GPT 4o seems to possess a more coherent and generalizable representation of mathematical principles, enabling it to maintain accuracy even as problem complexity increases. Its outstanding performance on word problems, achieving a 98% success rate compared to Meta AI's 4%, underscores its superior ability to translate natural language descriptions into mathematical formulations—a skill that requires a deeper conceptual understanding rather than mere pattern matching.

Claude Pro and Gemini Advanced exhibited strong, albeit not exceptional, mathematical reasoning, with overall success rates of 85.74% and 84.42%, respectively. Their specific strengths varied across different problem categories, suggesting differing emphasis in their training. Claude Pro showed notable strength in applying the product rule, with a 94% success rate, and implicit differentiation at 96%. Meanwhile, Gemini Advanced excelled with the power rule, achieving a 99% success rate, and performed well on conceptual problems related to increasing and decreasing intervals (70%, second only to Chat GPT 4o). These distinct performance profiles indicate that varied training approaches can lead to diverse strengths across mathematical domains, even when overall performance appears similar.

Copilot Pro's moderate performance, with a 76.30% overall success rate, revealed consistent limitations across most problem types, lacking both areas of excellence and catastrophic weaknesses. This pattern suggests that while Copilot Pro has a balanced mathematical understanding, it is less developed than the top three models in terms of reasoning capabilities. Its relatively uniform performance profile - with success rates typically in the 70-90% range across problem types - indicates a foundational level of mathematical knowledge, but it lacks the sophisticated reasoning skills needed for consistently high performance on complex problems.

Meta AI's notably lower performance, with a 56.75% overall success rate, highlighted fundamental limitations in mathematical reasoning that were evident across most categories. Although its performance was not uniform - achieving perfect scores on basic differentiation problems (100% for limit processes and the chain rule) while scoring only 4% on word problems - this extreme variability suggests that Meta AI may have access to a substantial amount of mathematical training data. However, it lacks the integrative reasoning capabilities necessary to apply this knowledge in complex or novel contexts. The stark contrast between its procedural competence and conceptual limitations offers valuable insights into the distinct components of mathematical reasoning that current AI systems must develop.

### 4.2 Problem Difficulty and Mathematical Reasoning Requirements

The varying performance across different problem types provides important insights into the relative difficulty of various calculus concepts for large language models (LLMs) and the distinct reasoning skills they require. The perfect performance of all models on limit-based differentiation (Problem 1) and chain rule application (Problem 5) indicates that procedural differentiation is effectively encoded in current LLM architectures. Although these problem types may seem conceptually complex to human learners, they follow algorithmic patterns that align well with the statistical pattern-matching capabilities of LLMs.

In contrast, the overall poor performance on increasing/decreasing interval analysis (Problem 8), which had a success rate of 49.0%, reveals a significant limitation in connecting derivative values to function behavior - an understanding that goes beyond procedural differentiation. This task requires interpreting the mathematical significance of sign changes rather than merely computing derivatives, highlighting a fundamental challenge in achieving genuine mathematical understanding in AI systems. The difficulty Claude Pro experienced with this type of problem (only a 23% success rate despite strong overall performance) suggests that specific conceptual connections may be lacking in otherwise capable models.

Word problems (Problem 10) presented the second-greatest challenge for all models, with an overall success rate of 58.8%. There were dramatic performance differences, with Chat GPT 4o achieving a success rate of 98% and Meta AI only 4%. The extreme difficulty these problems posed for some models demonstrates a fundamental limitation in translating natural language descriptions into mathematical formulations - an essential capability for applied mathematics. Solving word problems requires multi-step reasoning that integrates contextual understanding with mathematical principles, which may explain why models trained primarily on pattern matching within mathematical expressions struggle with this integration.

The varying performance on problems requiring algebraic manipulation after differentiation (particularly Problems 3, 4, 6, and 7) suggests that symbolic manipulation capabilities differ significantly across models. Meta AI's particular struggle with algebraic simplification, evident in its 35% success rate on product rule problems despite correctly applying the differentiation rule, highlights a disconnect between applying rules and subsequent mathematical processing. This finding aligns with research indicating that current LLMs may develop compartmentalized mathematical capabilities rather than integrated reasoning systems akin to human understanding [9], [10].

The cross-evaluation matrix revealed that problems generated by Claude Pro were consistently the most challenging, with an overall solution rate of 82.79%. This suggests that Claude Pro creates problems with particular mathematical nuance or complexity. The ability to generate challenging yet valid problems require a deeper mathematical understanding than merely solving pre-formulated problems, as it involves reasoning backward from desired properties to appropriate function structures. Notably, the fact that the difficulty of problem generation did not correlate directly with problem-solving performance - given that Meta AI generated the second-most difficult problems despite its poor solving performance - suggests that these capabilities depend on distinct aspects of mathematical representation within LLM architectures.

### 4.3 Algebraic Manipulation vs. Conceptual Understanding

A critical pattern emerging from the error analysis is the distinction between procedural differentiation, algebraic manipulation, and conceptual understanding - three distinct components of mathematical reasoning that are unevenly developed across current LLMs. All models demonstrated strong procedural differentiation capabilities, successfully applying differentiation rules in most cases. However, performance diverged significantly in the areas of algebraic manipulation and conceptual interpretation.

Errors in algebraic manipulation, especially in simplifying expressions, factoring polynomials, and handling complex fractions - accounted for the majority of mistakes across all models, with Meta AI showing particular weakness in this area (62% of its errors). This finding suggests that while symbolic manipulation capabilities have improved compared to earlier AI systems, they still represent a significant challenge for current LLMs. The frequent occurrence of algebraic errors highlights the difference between recognizing which differentiation rule to apply (a task well-suited to neural networks) and performing subsequent mathematical operations (a task that may require different architectural approaches).

Errors in conceptual understanding were less common but revealed more fundamental limitations in mathematical reasoning. These errors often involved misinterpreting the relationship between derivatives and function behavior, failing to recognize when optimization criteria were met, or incorrectly classifying critical points. The prevalence of these errors in Problems 8 (increasing/decreasing intervals), 10 (optimization), and 12-13 (second derivative applications) suggest that linking mathematical operations to their conceptual meanings remains challenging for current LLMs.

The disparity between procedural competence and conceptual understanding observed across all models has significant implications for educational applications. Students utilizing these systems as learning aids may receive correct differentiation procedures but flawed conceptual explanations, which could reinforce procedural approaches to calculus at the expense of deeper understanding. This limitation is particularly concerning for problems requiring the interpretation of derivatives in real-world contexts, where a solid conceptual understanding is essential for meaningful application.

### 4.4 Implications of Cross-Model Evaluation Patterns

The cross-evaluation matrix uncovered interesting patterns regarding how different models performed on problems generated by various sources. Each model achieved its highest success rates on problems it generated itself, indicating that a model's problem-solving approach aligns best with its own style of problem formulation. This self-compatibility effect was most pronounced for Chat GPT 4o, which solved 98.1% of its own problems, while it was least evident for Meta AI, which solved only 58.2% of its own problems.

Interestingly, problems generated by Meta AI were generally solved more successfully by other models than by Meta AI itself. The average success rate across the other models was 75.4%, compared to Meta AI's 58.2% self-success rate. This counterintuitive finding suggests that Meta AI creates mathematically valid problems with standard structures but struggles to apply consistent reasoning in their solutions. This discrepancy may indicate a disconnect between its problem generation and problem-solving capabilities. Furthermore, problems generated by Claude Pro were found to be the most challenging across all models, with an overall solution rate of 82.79%. This suggests that Claude Pro generates problems with particular mathematical nuances. An examination of these problems revealed that Claude Pro frequently designed functions that required careful attention to algebraic details and edge cases, aspects that proved to be challenging for other models. This ability might reflect Anthropic's "constitutional AI" approach, which emphasizes careful reasoning and attention to detail.

Overall, the relative difficulty of problems generated by different models showed little correlation with the models' problem-solving capabilities. This suggests that problem generation and problem-solving are distinct aspects of mathematical reasoning. Problem generation involves constructing functions with specific derivative properties - an inverse task that requires a deeper conceptual understanding than straightforward differentiation. The fact that Claude Pro excelled at generating challenging problems despite not being the top solver suggests it may possess stronger inverse reasoning capabilities than the other models, an attribute that is not fully captured by simple problem-solving evaluations.

### 4.5 Educational Implications and Applications

The findings have several important implications for the use of large language models (LLMs) in calculus instruction. The varying performance across models and problem types highlights the need for careful evaluation when integrating these tools into educational settings. Educators should note that even the strongest model, Chat GPT 4o, does not achieve perfect accuracy, particularly when tackling more complex problem types.

The distinct strengths and weaknesses identified for each model suggest potential complementary roles in educational contexts. Chat GPT 4o's exceptional performance across all categories makes it well-suited as a general calculus learning aid, especially for word problems and applications that require translating between different contexts. Claude Pro's ability to generate challenging problems indicates its potential value as a practice problem generator, particularly given its strong performance on the product rule and implicit differentiation. Gemini Advanced's relative strength in analyzing increasing and decreasing intervals, compared to other models (except for Chat GPT 4o), may make it particularly useful for conceptual explanations in this area. A common weakness observed across most models is their difficulty in connecting derivatives to functional behavior (as seen in Problem 8), underscoring a critical area where human instruction remains essential. Educators should exercise caution when relying on LLMs for conceptual explanations about how derivatives relate to function properties, as current AI systems demonstrate significant limitations in this domain. Moreover, the frequent algebraic errors found even in top-performing models indicate that human verification is necessary for complex calculus problems.

The ability of these models to generate valid practice problems represents a potentially valuable educational resource, albeit with important caveats. Both Claude Pro and Chat GPT 4o demonstrated sophisticated problem generation capabilities, producing mathematically sound problems with appropriate complexity. However, the limited success of Meta AI and Copilot Pro in generating advanced problems (Types 11-13) suggests that only select models should be relied upon for this purpose, emphasizing the need for human oversight. Furthermore, the observed performance patterns provide insights into potential pedagogical applications beyond direct problem-solving. The cross-evaluation matrix revealed that problems generated by Claude Pro were particularly challenging for all models, suggesting this approach could be beneficial for creating advanced practice materials that target specific mathematical skills. Additionally, the finding that Meta AI struggled significantly with word problems indicates that these types of problems require distinct reasoning capabilities that educational AI systems should specifically aim to address.

**4.6 Limitations in Current LLM Mathematical Capabilities**

Despite the impressive performance of top models, the results reveal several fundamental limitations in the mathematical reasoning capabilities of current large language models (LLMs).

First, the frequent occurrence of algebraic manipulation errors, even in models that correctly apply differentiation rules, indicate limitations in symbolic processing. These errors typically involve simplifying expressions, factoring polynomials, or manipulating complex fractions - tasks that require precise symbolic manipulation rather than mere pattern recognition.

Second, all models exhibited varying degrees of conceptual limitations, particularly in problems that required the interpretation of derivatives. The relatively poor performance on increasing/decreasing interval analysis (Problem 8, with an overall success rate of only 49.0%) reveals difficulty in connecting symbolic derivatives to their geometric and functional meanings. Likewise, the challenges models faced with optimization word problems (Problem 10, particularly for Meta AI with just 4% success) suggest limitations in translating contextual descriptions into mathematical formulations - a capability that is essential for applied mathematics.

Third, the cross-evaluation results highlighted limitations in consistency and generalization. Models often performed significantly better on problems that resembled their training examples than on those with unfamiliar structures or edge cases. This pattern suggests that current LLMs may develop fragile mathematical capabilities tied to specific problem formulations rather than a generalizable mathematical understanding. The finding that problems generated by Claude Pro were particularly challenging for all models emphasizes the limitations in handling mathematical structures that deviate from standard patterns.

Fourth, the difficulties models encountered in generating valid problems with specific properties (especially for types 11-13) highlight limitations in inverse mathematical reasoning. Creating functions with predetermined derivative characteristics requires reasoning backward from desired properties to appropriate function structures - a capability distinct from forward differentiation. The observation that even top-performing models needed iterative guidance to generate valid advanced problems points to fundamental limitations in this area of mathematical reasoning.

Finally, the significant performance gap between the best and worst models (94.71% for Chat GPT 4o versus 56.75% for Meta AI) indicates that mathematical reasoning capabilities vary dramatically across current LLM implementations. This variability suggests that mathematical competence is highly sensitive to specific architectural choices, training methodologies, and data selection - highlighting both the challenges and opportunities for further advancement in this field.

## 5 Conclusion

This comprehensive evaluation of five leading large language models (LLMs) on calculus differentiation problems reveals significant variability in their mathematical problem-solving capabilities. The performance hierarchy established through rigorous testing demonstrates substantial differences in how effectively these systems handle mathematical reasoning tasks. Chat GPT 4o showed the strongest overall performance with a 94.71% success rate, followed by Claude Pro (85.74%), Gemini Advanced (84.42%), Copilot Pro (76.30%), and Meta AI (56.75%).

The differential performance across problem types provides insight into the specific strengths and limitations of current LLM technologies. All models demonstrated mastery of basic differentiation techniques, achieving perfect scores on limit-based differentiation (Problem 1) and chain rule applications (Problem 5). This uniform excellence suggests that procedural differentiation aligns well with the pattern recognition capabilities of large language models, likely due to the algorithmic nature of these tasks and their prominent representation in training data.

However, as problems required greater conceptual understanding or multi-step reasoning, performance diverged significantly. Problems involving increasing and decreasing intervals (Problem 8) and optimization word problems (Problem 10) proved particularly challenging, with overall success rates of only 49.0% and 58.8%, respectively. These problem types require interpreting the mathematical meaning of derivatives rather than merely computing them - precisely the conceptual understanding that appears to be most challenging for current AI systems to develop. The substantial performance gap on these problems reveals a fundamental limitation in connecting symbolic manipulations to their mathematical meanings.

Error pattern analysis revealed a consistent hierarchy of difficulties across all models. Procedural differentiation errors were relatively uncommon, suggesting that the mechanical application of differentiation rules has been effectively encoded in current LLMs. In contrast, errors in algebraic manipulation - particularly in simplifying expressions, factoring polynomials, and handling complex fractions - occurred frequently, especially for Meta AI (62% of its errors) and Copilot Pro (41%). Most concerning from an educational perspective were the errors in conceptual understanding, which revealed fundamental limitations in interpreting derivative meanings and applying them in context.

The cross-evaluation matrix provided valuable insights into the relative difficulty of problems generated by different models and how these challenges affected performance. Problems generated by Claude Pro proved to be the most challenging across all models, with an overall success rate of 82.79%, suggesting that this model creates problems with particular mathematical subtlety. The finding that problem-generation capability did not correlate directly with problem-solving performance indicates that these are distinct aspects of mathematical reasoning, with generation potentially requiring deeper conceptual understanding than solving.

For educators and students, these findings suggest that LLMs can serve as valuable supplementary tools for calculus education but should not be relied upon uncritically. Even the best-performing model, Chat GPT 4o, fell short of perfect accuracy, with performance decreasing for more complex problem types. The specific strengths and weaknesses identified for each model suggest potential complementary roles in educational settings - Chat GPT 4o for general assistance, Claude Pro for generating challenging practice problems, and human instructors for providing conceptual explanations of derivative meanings and interpretations.

The methodology developed for this study - systematic evaluation across problem types with cross-model problem generation - provides a valuable framework for assessing mathematical reasoning in AI systems. By examining performance across a range of calculus concepts rather than relying solely on aggregate metrics, this approach offers nuanced insight into specific strengths and limitations. The cross-evaluation matrix reveals patterns in how models interact with problems generated from different sources, providing a more comprehensive understanding of mathematical reasoning capabilities than traditional benchmarking can achieve.

Several limitations of the current study should be acknowledged. First, the evaluation focused exclusively on differentiation problems, which represent only one domain within calculus. Performance on these problems may not generalize to other areas, such as integration, series, or differential equations, which involve different patterns of mathematical reasoning. Second, the evaluation was conducted at a specific point in time with the then-current versions of each model. Given the rapid pace of LLM development, these results may not reflect the capabilities of newer versions. Lastly, the standardized prompting approach, while necessary for fair comparison, may not accurately represent how students and educators interact with these models in practice, where iterative clarification is common.

As AI systems continue to advance, the distinction between procedural mathematics and conceptual understanding is likely to become increasingly important. Current LLMs demonstrate impressive capabilities in procedural tasks but exhibit significant limitations in conceptual understanding, mirroring challenges in human mathematics education. Understanding this distinction and developing systems that effectively bridge represent a critical frontier for both AI research and educational practice.

In conclusion, this study provides a comprehensive benchmark of current LLM capabilities in calculus problem-solving, revealing both remarkable achievements and significant limitations. While these models have developed sophisticated procedural capabilities that can support mathematics education in valuable ways, they fall short of the integrated conceptual understanding that characterizes human mathematical reasoning. As these technologies continue to evolve, careful evaluation of their specific strengths and limitations will remain essential for appropriate educational integration and targeted improvement. The methodology and findings presented here provide a solid foundation for this ongoing exploration.